# FPGA Implementation of the CAR Model of the Cochlea

Chetan Singh Thakur, Tara Julia Hamilton, Jonathan Tapson and André van Schaik
The MARCS Institute, University of Western Sydney, Kingswood 2751, NSW, Australia
Email: C.SinghThakur@uws.edu.au, T.Hamilton@uws.edu.au, J.Tapson@uws.edu.au,
A.vanSchaik@uws.edu.au

Richard F. Lyon
Google, Inc., Mountain View,
CA 94043 USA
Email: dicklyon@acm.org

*Abstract*—The front end of the human auditory system, the cochlea, converts sound signals from the outside world into neural impulses transmitted along the auditory pathway for further processing. The cochlea senses and separates sound in a nonlinear active fashion, exhibiting remarkable sensitivity and frequency discrimination. Although several electronic models of the cochlea have been proposed and implemented, none of these are able to reproduce all the characteristics of the cochlea, including large dynamic range, large gain and sharp tuning at low sound levels, and low gain and broad tuning at intense sound levels. Here, we implement the 'Cascade of Asymmetric Resonators' (CAR) model of the cochlea on an FPGA. CAR represents the basilar membrane filter in the 'Cascade of Asymmetric Resonators with Fast-Acting Compression' (CAR-FAC) cochlear model. CAR-FAC is a neuromorphic model of hearing based on a pole-zero filter cascade model of auditory filtering. It uses simple nonlinear extensions of conventional digital filter stages that are well suited to FPGA implementations, so that we are able to implement up to 1224 cochlear sections on Virtex-6 FPGA to process sound data in real time. The FPGA implementation of the electronic cochlea described here may be used as a front-end sound analyser for various machine-hearing applications.

## I. Introduction

The efficiency of the human auditory system in perceiving sound is superior to any machine hearing application. Its front end, the cochlea, is a complex three-dimensional fluid-filled structure that converts mechanical vibrations induced by an input sound signal into neuronal spikes on the auditory nerve [1]. For decades, efforts have been made to engineer a hearing machine that can emulate the function and efficiency of the cochlea and the human auditory system. Despite tremendous progress, the existing machine hearing applications are not capable of performing as well as their biological counterpart.

In human hearing, sound is collected in the outer ear, vibrates the eardrum and is transmitted via the middle ear bones to deliver hydrodynamic waves through the oval window of the inner ear's cochlea. These waves are coupled to mechanical vibrations of the basilar membrane (BM) in the fluid-filled cochlea. The stiffness of the BM declines with distance from the oval window, thus allowing it to act as a frequency spectrum analyser by vibrating at specific cochlear locations depending on the input frequency.

The motion of the BM is transduced by the inner hair cells into neural signals along the auditory nerve fibres. The cochlea thus exploits the physics of wave propagation through a non-uniform medium and employs sophisticated neural machinery to achieve remarkable acoustic sensitivity, high frequency selectivity and nonlinear compression in processing sound signals [1].

Several electronic models of the cochlea have been implemented in both digital and analogue VLSI technology since the first model was proposed by Lyon and Mead [2]. These models have found applications in audio signal processing systems for speech recognition, pitch detection and spatial localisation [3], [4]. However, the tremendous computational burden of the digital models limits their ability to run in real-time at low cost and low power. Some of these limitations have been addressed by using analogue VLSI models [5], [6]. Although analogue VLSI implementation offers the advantages of high speed and low power consumption, field programmable gate array (FPGA)-based models are superior in terms of shorter design and fabrication times, higher accuracy, and a simpler computer interface. With improvements in FPGA technology, it is now possible to develop large scale neuromorphic systems on a single FPGA chip [7].

In this paper, we describe an FPGA implementation of the 'Cascade of Asymmetric Resonators' (CAR) model of the BM as a first step in implementing the 'Cascade of Asymmetric Resonators with Fast-Acting Compression' (CAR-FAC) model of the cochlea. CAR-FAC is a neuromorphic model of hearing that incorporates recent findings on cochlear wave mechanics [8]. The model employs a filter-cascade approach that closely mimics the way sound information propagates as travelling waves in the human cochlea. It runs fast as its computational load is equivalent to that of a second-order filter per output channel, and is an efficient alternative to the more conventional parallel filter bank approach [9]. To achieve accuracy similar to the human cochlea, which can detect frequencies in the range of 20 Hz to 20 kHz, a model requires a large number of filter channels. Here, we show that an electronic cochlea with 1224 filter sections can be implemented on an FPGA (Virtex-6) to process real-time sound input, which is difficult to achieve at such a high resolution via software implementation. Our work demonstrates the capability of the CAR model to process sound at high resolution in real-time. The FPGA implementation of the electronic cochlea described here may be used as a front-end sound analyser for various machine-hearing applications.

This work has been supported by the Australian Research Council Grant DP0881219.

## II. OVERVIEW OF THE CAR-FAC MODEL OF THE COCHLEA

The CAR-FAC model is a dynamic digital version of the pole-zero filter cascade auditory filter model [10]. It closely approximates the physiological elements of the human cochlea (Fig. 1A) and mimics its qualitative behaviour. CAR-FAC comprises a cascade of asymmetric resonators (CAR) that models waves on the BM, inner and outer hair-cell (IHC and OHC) models, and a coupled automatic gain control (AGC) network that implements much of the compression part of the model. The OHCs provide dynamic nonlinearity or fast acting compression (FAC) in the CAR-FAC model. AGC functions as a feedback loop that controls the OHC. The IHCs connect the mechanical waves on the BM to neural signals on the auditory nerve. The IHC also plays a key role in the feedback loop that controls the adaptive gain and distortion in the mechanics. Fig. 1B shows the connections between the various elements of the CAR-FAC model.

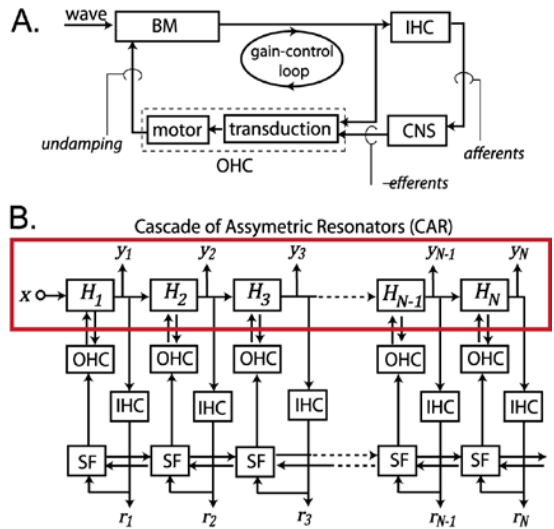

**Fig. 1. (A) Functional physiological elements of the cochlea.** The sound stimuli induce mechanical vibrations on the basilar membrane (BM). The inner hair cells (IHCs) are the transducers that sense sound-generated motion of the BM and deliver neural signals to the auditory central nervous system (CNS). The gain is provided by outer hair cells (OHCs) that provide active undamping, and the degree of undamping is controlled by efferent connections. Adapted from [12]. **(B) Architecture of the CAR-FAC model.** The CAR-FAC model includes the cascade of asymmetric resonators (CAR) which are quasi-linear transfer functions $H_1$ through $H_N$ that model BM motion. Fast-Acting Compression (FAC) is implemented by the OHC model integrated with the filter sections, and the coupled AGC (smoothing filters, SF) that controls the parameters of the filters via the OHCs. The IHCs introduce nonlinearity in the outputs of the CAR using sigmoidal or half-wave rectification function. The lateral interconnections of the smoothing filters allow a diffusion-like coupling across both space and time. Outputs from the CAR-FAC include BM motion ($y_1$ through $y_N$) and an estimate of average instantaneous rate on the auditory nerve, the neural activity pattern ($r_1$ through $r_N$). Adapted from [13].

The asymmetric resonators in the cascade of asymmetric resonators (CAR) are two-pole–two-zero filters. The number of filter sections and their coefficients are optimised to match a linearised model of the human cochlea. The pole frequencies are chosen to correspond to equal spacing along the place dimension of the cochlea, by using the Greenwood function for human cochlea [11]:

$$f = 165.4(10^{2.1x} - 1) \quad (1)$$

where, $x$ is normalised with respect to the length of the cochlea and varies from 0 at the apex to 1 at the basal end. Fig. 2 shows one such filter section, where $a_0$ and $c_0$ are functions of position, $x$, along the cochlea.

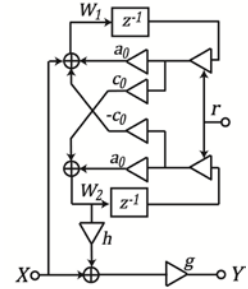

**Fig. 2. Biquadratic filter in the CAR model.** $X$ is the input signal, $Y$ is the output signal, and $W_1$ and $W_2$ are internal state variables.

The $a_0$ and $c_0$ parameters represent the analogue pole position in the zero-damping case. An explicit parameter '$r$' can be used to dynamically vary the pole and zero radius in the z plane (to vary the damping factor). '$r$' is controlled by OHC to control the gain.

$$a_0 = cos(\theta_R) = a/r \quad (2a)$$
$$c_0 = sin(\theta_R) = c/r \quad (2b)$$

where, $\theta_R$ is the normalised pole ringing frequency in radians per sample, or pole angle in the z plane. Using these parameters, the transfer function is:

$$Y/X = g[(z^2 + (-2a_0 + hc_0)rz + r^2)/(z^2 - 2a_0rz + r^2)] \quad (3)$$

The $h$ coefficient controls how far the zeros are from the pole frequency, and the $g$ coefficient is used to adjust the overall gain. As long as $h$ is small enough that the zeros remain complex, the zeros will be at the same radius $r$ as the poles. The condition for complex zeros becomes relevant for high-frequency channels, where $cos\,\theta_R < 0$. In that case:

$$h < [(2 + 2a_0)/c_0] \quad (4)$$

To get unity gain at DC, we can solve for g:

$$g = [(1 - 2a_0r + r^2)/(1 - (2a_0 - hc_0)r + r^2)] \quad (5)$$

The cascaded stages are combined to provide a family of filters at the output taps between the stages. The resulting filters may have high peak gains, depending on the stage damping parameters.

## III. DESIGN METHODOLOGY AND FPGA IMPLEMENTATION

Here, we will describe the FPGA implementation of the cascade of asymmetric resonators (CAR) that represent the BM filter in the CAR-FAC model of the cochlea. Each filter section is a linear two-pole–two-zero filter, also known as biquadratic filter section, described by the transfer function in Eq. 3. The impulse response of each filter is governed by different coefficients. Each location in the cochlear model is tuned for a different frequency as determined using the Greenwood map (Eq. 1), ranging from the highest frequency of 20.657 kHz (corresponding to $x = 1$) to 20 Hz (corresponding to $x = 0.023$). Here, $x$ varies from 0 to 1 from the apex of the cochlea to its basal end. The filter sections are cascaded from high to low resonant frequency similar to the BM. A sound input passes through the cascade of filter sections and excites a range of filters tuned near the corresponding frequency.

### A. Software Implementation

First, we simulated a software floating-point implementation of CAR in Python. The coefficients $a$, $c$, $h$ and $g$ for each filter section were calculated using Eqs. 2a, 2b, 4 and 5, respectively, with $r$ chosen as a free parameter. Next, we adapted the Python code for fixed-point implementation, and determined the word length of the input, output and internal variables required for FPGA implementation without losing accuracy.

### B. FPGA Implementation

Xilinx Virtex-6 (XC6VLX240T-1FF1156) FPGA was used to implement the CAR model. The sound input had a sample frequency of 48 kHz. Using a system clock frequency of 142 MHz, one physical CAR section requires 29 system clock cycles to generate the output, resulting in a latency of 203 ns for a single section.

As shown in the block diagram in Fig. 3, there exists a global state machine which determines the filter section to be processed at a particular time and controls the coefficients and data for that section. In the CAR core block, there are two parallel state machines which control and calculate internal variables W1 and W2, which further calculate the transfer function (Eq. 3). For each sample, the global state machine keeps track of the filter section number and controls the data flow of the CAR core by passing the required input data for each filter section to the CAR core block, which then processes the sound sample. The coefficients $a$, $c$, $g$ and $h$ for each filter section were calculated externally, and uploaded into the FPGA from a file. The filters use a delay element ($z^{-1}$ block in Fig. 2) that requires two internal variables, W1 and W2, to be stored for each filter section. After completion of the operation for one filter section, it asserts a 'done' signal and passes the output to the global state machine. All filter stages are cascaded, i.e., the output of one stage is input for the next stage. After completion of all filter stages, the global state machine asserts the 'done_sample' signal, which signifies that all the sections have been processed for a given input sample.

As the sound input data rate is 48 kHz (i.e., time period = 20.8 µs), all filter sections need to finish their operation in less than 20.8 µs. Given a system clock frequency of 142 MHz, and the fact that each filter section requires 29 clock cycles, it is possible to implement 102 filter sections, henceforth termed a CAR array, using time multiplexing of a single hardware filter. Given the size of a Virtex-6 FPGA, we were able to implement 12 CAR arrays for a total of 1224 cochlear sections. This introduces a latency of 250 µs at the final section, which is perfectly acceptable for the lowest frequency when simulating the biological cochlea. The device utilisation summary is presented in Table 1

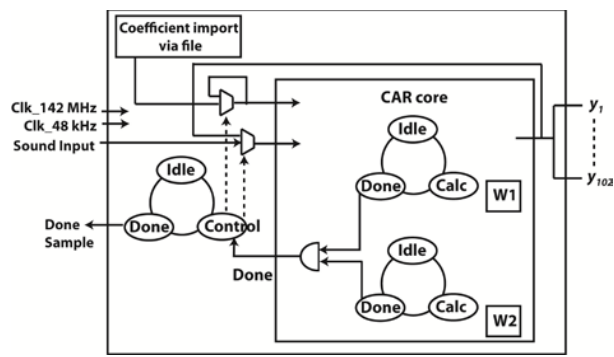

**Fig. 3.** Simplified architecture of a CAR array, which implements 100 filter sections with a CAR core using time multiplexing.

TABLE I. DEVICE UTILISATION SUMMARY

| Slice Logic Utilisation | Used | Available | Utilisation |
|---|---|---|---|
| Number of Slice Registers | 113,760 | 301,440 | 37% |
| Number of Slice LUTs | 136,957 | 150,720 | 90% |
| Number used as Memory | 3,005 | 58,400 | 5% |

## IV. RESULTS

We have implemented the CAR model in FPGA and compared the results with the software simulations of the model (Fig. 4). The impulse response and gain for 20 filter sections, measured using a maximum-length sequence as the input signal, are shown. The hardware implementation was verified by comparing Cadence NCSIM simulations with the results produced by the floating-point Python model.

## V. CONCLUSIONS

In this paper, we have shown that we can implement up to 1224 filter sections on a Xilinx Virtex-6 FPGA, due to low system complexity and ease of hardware implementation of the CAR model. Future work will aim to reduce the number of multipliers using resource allocation technique, so as to increase the number of filter sections on FPGA. Future work will also include implementation of IHC, OHC and AGC modules of the CAR-FAC model and integration of these elements with the filter sections. Furthermore, with the integration of all system elements, '*r*' will be variable to control damping dynamically. Nonetheless, the FPGA implementation of the electronic cochlea described here may be used as a front-end sound analyser for various machine hearing applications.

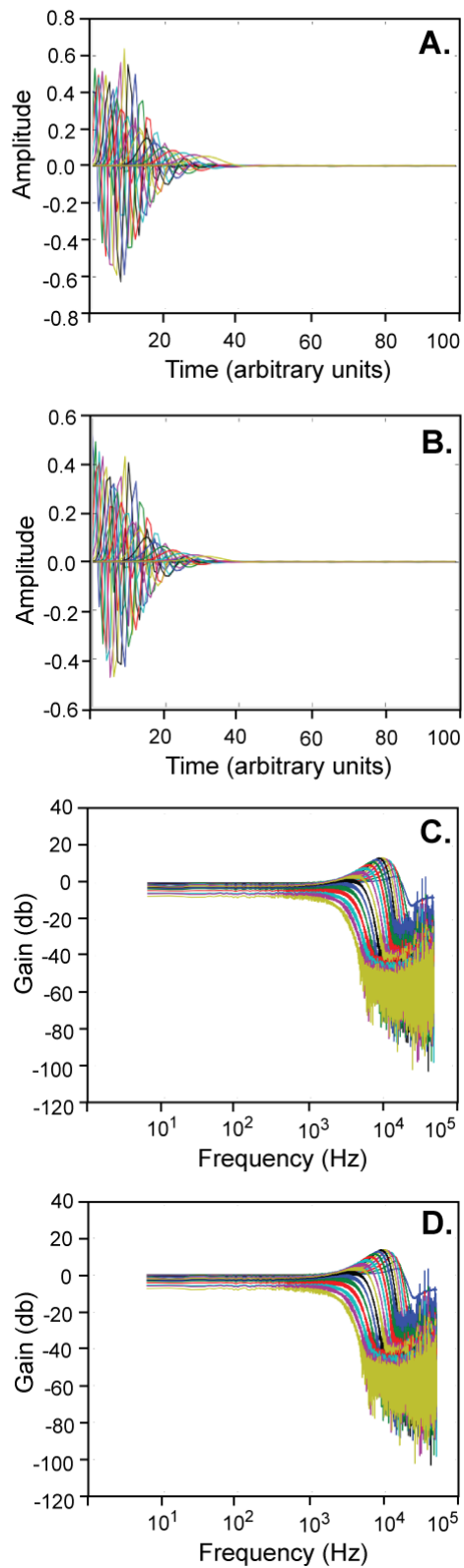

**Fig. 4. Implementation of CAR model in software and hardware (FPGA)**. Impulse response of (**A**) Software floating-point implementation and (**B**) Hardware fixed-point implementation. Frequency response of (**C**) Software floating-point implementation and (**D**) Hardware fixed-point implementation.